\documentclass{article}
\usepackage{iclr2027_conference,times}
\usepackage[T1]{fontenc}
\usepackage{graphicx}

\usepackage{amsmath,amssymb}
\usepackage{algorithm}
\usepackage{algorithmic}
\usepackage{booktabs,tabularx,multirow}
\newcolumntype{C}{>{\centering\arraybackslash}X}
\newcommand{\tablehead}[1]{%
  {\renewcommand{\arraystretch}{0.9}%
   \begin{tabular}[c]{@{}c@{}}#1\end{tabular}}}
\newcommand{\tablemethod}[2]{%
  {\renewcommand{\arraystretch}{0.9}%
   \begin{tabular}[c]{@{}l@{}}#1\\[-1pt]{\scriptsize #2}\end{tabular}}}

\usepackage{flafter}
\usepackage{placeins}
\usepackage[hidelinks]{hyperref}
\usepackage{url}

\newcommand{\ind}{\mathbf{1}}
\newcommand{\supp}{\mathsf{S}}
\newcommand{\contra}{\mathsf{C}}
\newcommand{\unk}{\mathsf{U}}
\DeclareMathOperator{\pa}{pa}
\DeclareMathOperator{\anc}{anc}

\title{SCOPE: Selective Certification\\
of Program Edits\\
for Embodied Heuristic Learning}
\author{Junyang Shu$^{1}$ \quad Bingqing Wei$^{1}$ \quad Jishuo Li$^{1}$\\
\textbf{Zheqi He$^{2}$ \quad Yongtao Wang$^{1,3}$}\\[3pt]
\normalfont $^{1}$\,Wangxuan Institute of Computer Technology, Peking University, China\\
\normalfont $^{2}$\,Beijing Academy of Artificial Intelligence\\
\normalfont $^{3}$\,VGI Labs}

\iclrfinalcopy
\hypersetup{
  pdftitle={SCOPE: Selective Certification of Program Edits for Embodied Heuristic Learning},
  pdfauthor={Junyang Shu, Bingqing Wei, Jishuo Li, Zheqi He, Yongtao Wang}
}

\begin{document}
\maketitle
% The preprint does not carry a conference publication or review header.
\fancyhead{}
\renewcommand{\headrulewidth}{0pt}
\begin{abstract}
Embodied heuristic learning must balance strategic adaptation with the
preservation of previously successful behavior. Local repairs may leave
structural failures unresolved, whereas broader revisions can introduce
regressions that are costly to detect through repeated environment interaction.
We introduce SCOPE (Selective Certification Of Program Edits), which uses
failed and successful executions as complementary evidence for program revision.
Frontier-guided program search annotates a shared diagnostic graph with
execution evidence to distinguish supported progress from unresolved checks.
Selective behavioral certification
uses stateful API replay to restore recorded responses and mutations, checking
whether edits preserve selected historical successes. Diagnosis and certification feedback,
supplemented by sparse environment diagnostics, guide consecutive repairs
across configurations. SCOPE outperforms the strongest baseline
in both simulation and real-world tasks. On LIBERO-PRO and Robosuite,
it achieves mean success rates of 80.66\% and 84.48\%, with respective gains
of 6.51 and 3.48 percentage points.
Our code is available at \url{https://github.com/VDIGPKU/SCOPE}.
\end{abstract}

\section{Introduction}
\label{sec:introduction}

Robot control programs provide an explicit representation of how perception,
decisions, and actions compose into task behavior. Language models can
generate these programs from natural-language instructions, using control
flow and API calls to express policies and situated task
plans~\citep{liang2023code, singh2023progprompt}. Environment feedback then
provides a basis for adapting behavior when execution departs from the
plan~\citep{huang2023inner}. This supports embodied heuristic learning,
in which interaction improves task-specific decision rules while
language-model parameters remain fixed. The resulting program must work across
configurations, so learning requires more than solving the latest observed
failure.

Execution-driven systems build on this approach through iterative program search.
CaP-X studies coding agents that compose robot APIs and revise programs
through interactive execution, while RHO optimizes policy repositories
using execution feedback and validation-based candidate
selection~\citep{fu2026capx, elmaaroufi2026rho}. Reusable knowledge provides
another source of improvement. Playful Agentic Robot Learning acquires code
skills through self-directed exploration, and ASPIRE combines multimodal
execution traces, validated repair knowledge, and evolutionary program
search~\citep{zhang2026playful, lu2026aspire}. These approaches establish
execution as both a source of learning evidence and a means of evaluating
candidate programs. As revisions accumulate, the effects of each edit on
previously successful behavior become an additional concern.

The difficulty is that a failure reveals where execution broke down, but
the required change may extend beyond that point. A failed lift, for
example, may originate in an unstable grasp, requiring the grasping stage
to be redesigned rather than the lifting height to be adjusted. Local
repairs can leave such structural failures unresolved, whereas broader
revisions can alter decisions that supported success in other
configurations. Success on the current configuration provides no direct
evidence about those earlier behaviors. Repeating environment execution
across historical successes can assess the change, but spends interaction
budget on checking old behavior as well as discovering new behavior.
Effective search therefore needs evidence to guide the extent of a repair
and assess its consequences for prior successes.

Our central insight is that failed and successful executions provide
complementary evidence for program revision. Failed traces guide diagnosis
and revision planning, while successful traces can be replayed to assess
whether earlier behavior is preserved after an edit. By executing a revised
program against the exchanges recorded in a successful trace, we can check
whether it makes the same API requests and consumes the full trace with
normal termination.
Under faithful recording and deterministic execution, with all relevant
external interactions captured at the API boundary, a complete match
certifies success on the recorded instance. A divergence instead identifies
where behavior changes and provides a target for further environment
diagnosis. This turns historical interaction into feedback for subsequent
program edits.

We introduce SCOPE (Selective Certification Of Program Edits), which couples
frontier-guided program search with this selective certification feedback.
SCOPE first constructs a shared diagnostic graph whose checks describe
observable task progress and whose edges encode prerequisites. Execution
evidence annotates the graph to identify a failure frontier consisting of
unresolved checks with supported prerequisites. This frontier and its dependencies
guide repair planning, allowing the model to select among local repair,
stage redesign, and strategy regeneration. Revisions accumulate in a shared
working program across development configurations.

SCOPE accepts the working program whenever it succeeds on a configuration.
If this changes the accepted source, SCOPE
samples historical successes for stateful full-trace API replay.
Replay preserves recorded responses and public-state
mutations, allowing later program decisions to depend on the same effects
as in the original execution. For selected configurations that remain
uncertified, sparse environment diagnostics provide fresh outcomes and
failure evidence. These observations feed back into the same repair loop.
Certification thus informs subsequent search through evidence of preserved
behavior and changed interactions.

We evaluate the frozen programs on held-out configurations in
LIBERO-PRO~\citep{zhou2025liberopro} and
Robosuite~\citep{zhu2020robosuite}. SCOPE achieves mean success rates of
80.66\% and 84.48\%, exceeding the strongest baseline on each benchmark
by 6.51 and 3.48 percentage points, respectively. Component ablations
support the roles of failure-frontier guidance, certification feedback,
and explicit revision-level planning. Across the reported settings,
full-trace replay takes 4.62--6.59\% of the corresponding original
environment execution time, demonstrating the low cost of obtaining
behavioral evidence from recorded interactions.

Our contributions are:
\begin{itemize}
    \item A program-search framework that uses execution evidence to
    localize failure frontiers and plan repairs at different revision
    levels, integrating historical-success feedback into consecutive
    repair.
    \item A selective certification mechanism based on stateful full-trace
    API replay, with an instance-specific success guarantee under the
    stated execution assumptions and sparse diagnostics for further repair.
    \item An evaluation on LIBERO-PRO and Robosuite showing improved
    mean held-out success over the strongest baseline on each benchmark,
    together with real-world experiments demonstrating strong performance
    on tasks requiring base--arm coordination, bimanual coordination,
    and fine manipulation.
\end{itemize}

\section{Preliminary}
\label{sec:preliminary}

We study \emph{embodied heuristic learning}, the refinement of task-specific
decision rules through interaction. Control programs encode these heuristics through
perception, conditional decisions, and actions.
Given a task description $d$ and public observation and control APIs,
$P\in\mathcal P$ maps observation histories to API calls. An execution
instance $\xi=(x_0,\omega,\kappa)$ specifies initial state and program
inputs $x_0$, exogenous randomness $\omega$, and API, runtime, and evaluator
semantics $\kappa$. Let $Y(P;\xi)\in\{0,1\}$ indicate complete task success
with normal termination. For the task distribution $\mathcal D_d$, we maximize
\begin{equation}
    J(P)=\mathbb E_{\xi\sim\mathcal D_d}\bigl[Y(P;\xi)\bigr],
    \label{eq:objective}
\end{equation}
using finite development reset configurations $\mathcal S$ and
per-configuration repair limits. Execution $e$ on $s_e\in\mathcal S$
produces recorded interactions $\tau_e$. Repeated resets have their own
observations and outcomes. Heuristic learning updates the program and memory
while keeping language-model parameters fixed.

SCOPE couples
\emph{frontier-guided search} with \emph{selective certification feedback}.
The shared working program accumulates revisions across configurations.
On each configuration, we execute it before repair, using observed failures
to guide consecutive revisions and successful executions to update the
accepted program. Replay and sparse diagnostics supply additional repair evidence.
At search trial $t$, SCOPE maintains the state
$\bigl(P_t^{\mathrm{acc}},P_t^{\mathrm{work}},
G,\mathcal A_t,\mathcal M_t\bigr)$, comprising the accepted program, a working program
shared across configurations, a diagnostic graph, an execution archive, and a
historical success bank.
Both programs start from a generated $P_0$. The archive and success bank
are populated as development configurations are visited. Entry trials,
repair candidates, and diagnostics are archived by role. Before a successful
execution is available, repairs use the current failure record alone,
allowing search to begin with $\mathcal M_0=\varnothing$.

The replay analysis assumes that all behavior-relevant external inputs
and effects cross the recorded API boundary, with program decisions and
environment responses deterministic given $\xi$. Records faithfully preserve
values, aliases, exceptions, and public-state mutations. Full matches cover
all external events and normal termination. The fixed evaluator determines
the outcome from the resulting state.

\section{Method}
\label{sec:method}

\begin{figure}[tb]
    \centering
    \includegraphics[width=\linewidth]{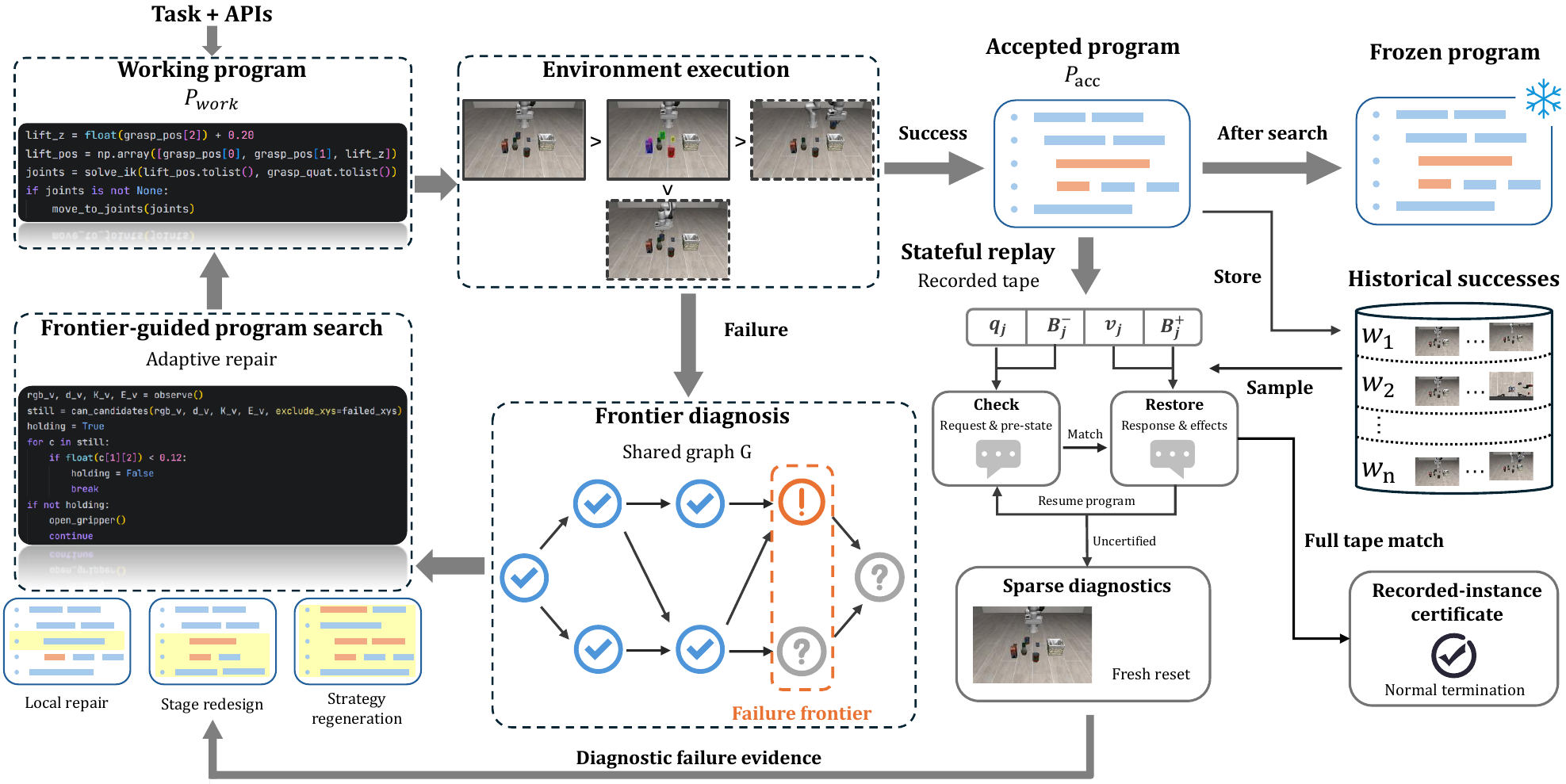}
    \caption{\textbf{Overview of SCOPE.} Execution evidence annotates a shared
    diagnostic graph to guide local repair, stage redesign, or strategy
    regeneration. Success updates the accepted program and historical
    success bank. When the accepted source changes, selected historical
    tapes are replayed by a local executor.
    Full matches certify the recorded instances under the replay assumptions.
    Sparse environment diagnostics on uncertified configurations supply
    failure evidence for subsequent repair. The accepted program is frozen
    after search.}
    \label{fig:overview}
\end{figure}

SCOPE couples frontier-guided program search with selective certification
feedback, as illustrated in Figure~\ref{fig:overview}. Failed executions
guide repairs, while historical successes provide evidence about the
behavior of updated programs.

\subsection{Failure-Frontier Memory and Retrieval}
\label{sec:frontier-memory}

At the start of each task, a model plans a directed acyclic
\emph{diagnostic graph} $G=(V,E)$ from the task description, initial observations, and API
specification. The graph is shared across development configurations.
Nodes define observable checks, and edges
encode prerequisites. Checks in $V_{\mathrm{inv}}\subseteq V$ require
sustained support while dependent checks are established. The graph
provides the reference for assessing execution traces and identifying
supported progress and unresolved stages for repair.

A separate critic assesses each execution $e$ against this graph, labeling
each node $c_{e,v}\in\{\supp,\contra,\unk\}$
(supported, contradicted, or unknown), citing a window
$I_{e,v}=[\ell_{e,v},u_{e,v}]$ in sampled public frames and API traces.
Here, $\ell_{e,v}$ and $u_{e,v}$ denote the window's start and end execution steps, respectively.
The host verifies frame provenance and cited-step endpoints, retaining raw judgments.
In topological order, it retains $c_{e,v}=\supp$ only when
\begin{equation}
\begin{aligned}
    c_{e,u}&=\supp
        &&\forall u\in\pa(v),\\
    u_{e,u}&\leq\ell_{e,v}
        &&\forall u\in\pa(v)\setminus V_{\mathrm{inv}},\\
    c_{e,u}&=\supp,\quad I_{e,v}\subseteq I_{e,u}
        &&\forall u\in\anc(v)\cap V_{\mathrm{inv}}.
\end{aligned}
\label{eq:evidence-consistency}
\end{equation}
Here $\pa(v)$ and $\anc(v)$ denote parents and ancestors. Ordinary
prerequisites must precede the dependent window, while invariant ancestors
must support its entire duration. Missing evidence or violated constraints
change the label to $\unk$.

The supported nodes $S_e=\{v\in V:c_{e,v}=\supp\}$ form the execution's
supported prefix in $G$. The \emph{failure frontier} contains unresolved
checks with supported prerequisites:
\begin{equation}
    F_e=\bigl\{v\in V\setminus S_e:\pa(v)\subseteq S_e\bigr\}.
    \label{eq:frontier}
\end{equation}
The archive stores $F_e$ with the source, raw execution, outcome, raw and
filtered judgments, evidence windows, and failure hypotheses. Each execution
thus yields an annotation of the shared graph, linking supported progress
and unresolved stages to trace evidence. This guides the model in locating
failures and inspecting their prerequisites. Records without usable
assessments still supply raw observations, traces, and errors for repair.

For code generation, SCOPE retrieves the current failure record and, when
available, the latest successful execution record associated with the
accepted program $P_t^{\mathrm{acc}}$ from the task archive.
The failure supplies local diagnostic evidence, while the successful record
provides a reference for behavior achieved by the accepted program.
Before the first successful search trial, only the failure record is used.
Each record retains its own source, observations, trace, outcome, and available
diagnostic evidence. Edits use the shared working source, carrying accumulated
revisions across configurations.

\subsection{Adaptive Program Search}
\label{sec:adaptive-search}

On its first main-search visit, each configuration is tested with the shared
working program on a fresh reset. Success ends the visit. After failure, the critic
annotates the trace against $G$, and the model revises the program using
this evidence. The next execution on the same configuration validates the
revision and, if it fails, supplies the diagnosis for another repair.
Each configuration allows at most three main-search repairs.
Replay feedback can trigger at most one additional follow-up repair per
configuration.

To plan each repair, the model forms a diagnosis from the shared working source,
the current execution's annotated graph, and the retrieved records.
Existing full-trace feedback from successful witnesses and diagnostic
outcomes also inform repairs (Section~\ref{sec:certification-feedback}).
Its repair plan states a failure hypothesis, cites execution
evidence, identifies decisions and code regions to change, and describes
the expected observable effect. Usable frontiers and dependencies help
localize the cause. Otherwise, planning uses raw observations, traces, and
errors. The model reports diagnosis, plan, and revision level in one planning
request.

The model chooses among three revision levels.
\emph{Local repair} changes a localized condition, parameter, or code fragment.
\emph{Stage redesign} changes a task stage and, when needed, its prerequisites,
while \emph{strategy regeneration} reorganizes the overall control strategy.
For example, evidence linking failed lifting to the grasp supports stage
redesign. An edit's behavioral divergence probability bounds both potential gains and
regressions (Appendix~\ref{sec:behavioral-interference}). This motivates
broader revisions when local repairs are insufficient, with execution
evidence guiding the choice.

Local and stage repairs edit the shared working source, while
regeneration may restructure it or restart from the accepted source.
The model generates a complete candidate $P'$ following the repair plan.
The candidate becomes the shared working program for the next
execution on that configuration. The working source is otherwise unchanged.
We archive the plan and revision level with the source and execution
outcome to guide subsequent repairs.

The main queue visits development configurations in a fixed order. A FIFO
queue holds diagnostic follow-ups. We alternate between nonempty queues
at visit boundaries, completing consecutive repairs on a configuration
before switching. Main-search and follow-up repairs are counted separately
for each configuration, with counts retained on re-entry.

\subsection{Success-Triggered Updates and Certification Feedback}
\label{sec:certification-feedback}

At search trial $t$, the working program executes on a fresh reset of $s_t$,
yielding $y_t=Y(P_t^{\mathrm{work}};\xi_t)$. Every successful entry,
candidate, or follow-up trial accepts the program that was executed:
\begin{equation}
\begin{aligned}
    A_t&=y_t\ind[P_t^{\mathrm{work}}\ne P_t^{\mathrm{acc}}],\\
    P_{t+1}^{\mathrm{acc}}
       &=\begin{cases}
           P_t^{\mathrm{work}}, & y_t=1,\\
           P_t^{\mathrm{acc}}, & y_t=0.
         \end{cases}
\end{aligned}
\label{eq:acceptance}
\end{equation}
Here $A_t$ marks a change in the accepted source. Each successful search
trial also updates the execution record associated with the accepted
program for subsequent repair retrieval. All execution records remain
archived. Full-trace checks and diagnostics follow accepted source updates,
adding evidence for later repairs while leaving both programs unchanged.

The bank stores one \emph{historical success witness}
$w=(P_w,\xi_w,T_w)$ per retained
configuration $s_w$, with $Y(P_w;\xi_w)=1$. Every successful search trial
adds or replaces the witness for $s_t$ and retains others, including those with later
diagnostic failures. Current-program outcomes are recorded separately.

\begin{figure}[tb]
    \centering
    \includegraphics[width=\linewidth]{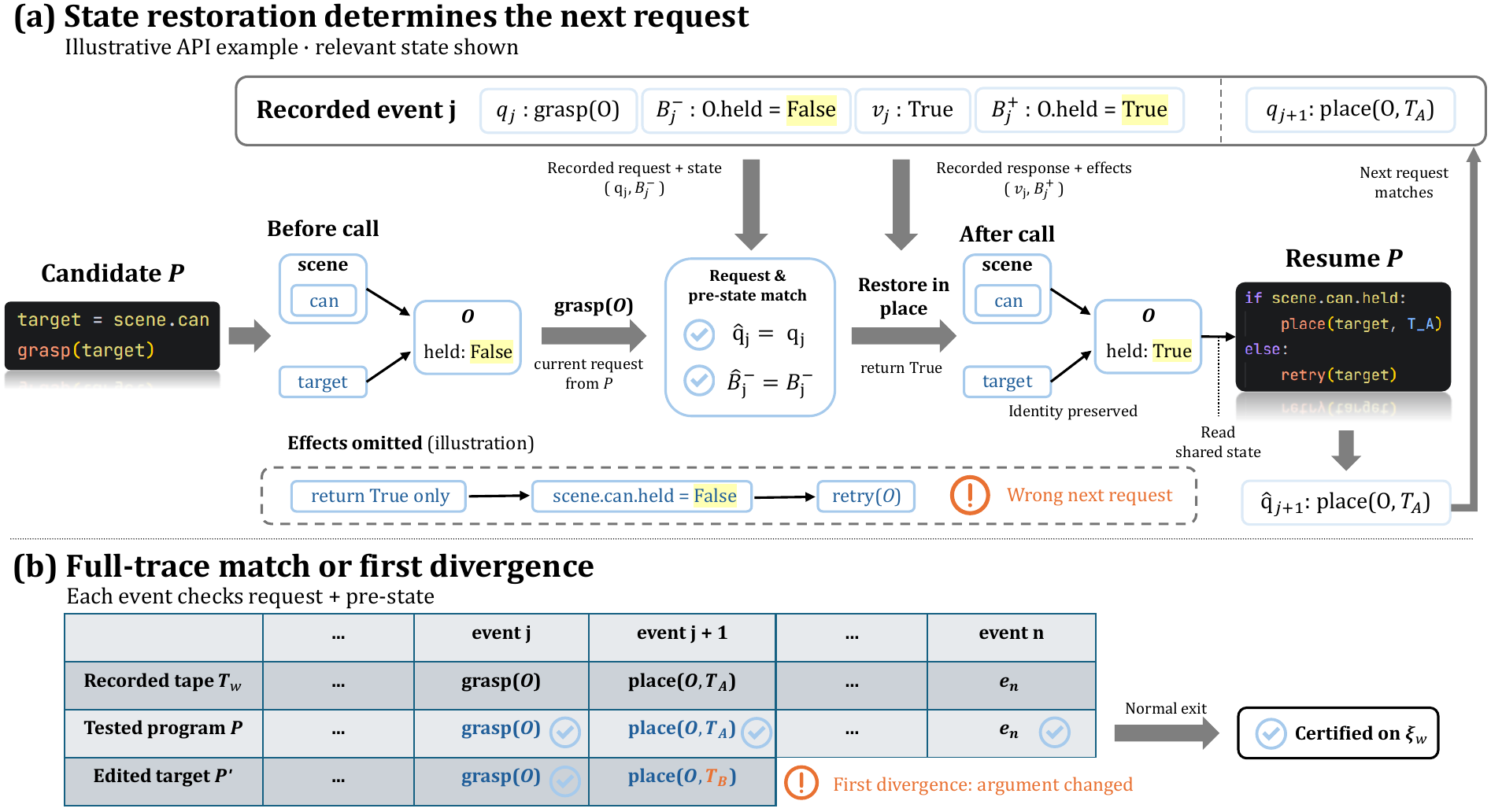}
    \caption{\textbf{Stateful full-trace replay.}
    \textbf{(a)} A local executor compares each API request and pre-call
    state with the tape, reproduces the recorded response and mutations
    while preserving shared references, and resumes the program.
    Omitting a mutation changes the next request in this illustrative
    example. \textbf{(b)} Matching every event, making no extra external
    calls, and terminating normally certifies success on the recorded
    instance $\xi_w$ under the replay assumptions. A changed argument stops
    replay at the first divergence, leaving the instance uncertified.}
    \label{fig:stateful-replay}
\end{figure}

SCOPE checks historical behavior through \emph{stateful full-trace replay},
which restores recorded API effects to preserve the state read by subsequent
program decisions. Figure~\ref{fig:stateful-replay} illustrates this mechanism
with a grasp-and-place example.
Each recorded execution $e$ supplies a stateful API tape
\begin{equation}
    T_e=\bigl((q_{e,j},B^-_{e,j},v_{e,j},B^+_{e,j})\bigr)_{j=1}^{n_e},
    \label{eq:effect-tape}
\end{equation}
where $q_{e,j}$ contains the API name and arguments, $v_{e,j}$ is the
return value or exception, and $B^-_{e,j},B^+_{e,j}$ are public object states
before and after the call. A canonical encoding preserves values, shared
references, array views, and API mutations to objects retained by the program.
A local executor runs the tested program from its entry point, comparing
each request and pre-call state with $(q_{e,j},B^-_{e,j})$. Matches apply
recorded mutations and reproduce $v_{e,j}$. Divergence stops replay.
A full-tape match also requires normal termination.

After an accepted source update ($A_t=1$), we sample up to $K=3$
success witnesses uniformly without replacement
from $\mathcal M_t$. This gives a certification set
$\mathcal W_t\subseteq\mathcal M_t$ that probes behavior preservation across
the historical success bank.
Let $C_t(P,w)=1$ denote a full-tape match with normal termination,
and $C_t(P,w)=0$ otherwise. Under the assumptions in
Section~\ref{sec:preliminary}, $C_t(P,w)=1$ certifies success on the
recorded instance $\xi_w$. Selected but uncertified witnesses are
\begin{equation}
    \mathcal U_t(P)=\{w\in\mathcal W_t:C_t(P,w)=0\}.
    \label{eq:unresolved-witnesses}
\end{equation}
Feedback distinguishes the first divergent interaction from unavailable
replay. Divergence indicates changed behavior, not task failure. Each record
identifies the tested program and witness so later repairs can locate
affected decisions.

After each accepted source update ($A_t=1$), \emph{sparse environment diagnostics}
execute the accepted program on at most $k_{\mathrm{diag}}$ configurations
from $\mathcal U_t(P_t^{\mathrm{work}})$ in the sampled order.
We select configurations whose follow-up repair remains unused and that
have no pending follow-up. A failed diagnostic enqueues the configuration
with source, trace, and available frontier evidence.

A follow-up repairs directly from the diagnostic failure if its source is
still the shared working program, using one execution to validate
the revision. Otherwise, it first executes the current program. Success
ends the visit, while failure prompts one repair and its validating trial.
Follow-ups use the same diagnosis, retrieval, and update rules as main
search. Fresh resets describe their own instances, which may differ from
$\xi_w$.
Replay and diagnostic counts are bounded per accepted source update.
Appendix~\ref{sec:certification-analysis} gives the certification proof
and cost analysis.
After completing search and servicing pending follow-ups, we freeze the
accepted program.

\section{Experiments}
\label{sec:experiments}

Our experiments address three questions:
(1) Does SCOPE improve the generalization of learned control programs to
held-out configurations?
(2) How do certification feedback, failure-frontier guidance, and
revision-level planning contribute to performance?
(3) How often does replay certify historical successes, and what
computational costs do certification and diagnostics incur?
We also evaluate frozen programs on three physical manipulation tasks
and examine how execution feedback changes a real-world control strategy.

\subsection{Experimental Setup}
\label{sec:experimental-setup}

We evaluate LIBERO-PRO~\citep{zhou2025liberopro}, an extension of
LIBERO~\citep{liu2023libero}, and Robosuite~\citep{zhu2020robosuite}.
LIBERO-PRO covers Object, Goal, and Spatial suites under position swaps
(Pos) and task changes (Task), totaling 60 task--setting pairs. Robosuite
covers seven manipulation tasks. Our primary comparison uses
GLM-5.3-Flash for both SCOPE and ASPIRE~\citep{lu2026aspire}, alongside
other program-based and vision-language-action baselines.

Programs are generated, repaired, and selected on development configurations,
then frozen for held-out evaluation without further repair.
Our simulation results report task-averaged success as mean $\pm$ sample
standard deviation over three independent learning and evaluation runs.
For the other comparison methods, we report point estimates under their
original protocols. Seed splits, implementation details, search budgets,
and baseline sources appear in Appendix~\ref{sec:experimental-details}.

\subsection{Benchmark Results}
\label{sec:benchmark-results}

On LIBERO-PRO, SCOPE achieves $80.66\,\pm\,1.46\%$ overall success
(Table~\ref{tab:libero-summary}), improving over ASPIRE with the same
GLM-5.3-Flash backbone by 6.51 percentage points and leading all six
suite--setting combinations. The largest gains over this baseline occur
on Goal--Task and Spatial--Pos, where success rises from 48.80\% to
61.87\% and from 54.47\% to 65.40\%, respectively.
The gains therefore extend beyond the Object suite, where both methods
already exceed 95\% in each setting. Individual task results appear in
Appendix~\ref{sec:libero-task-results}.

% Generated by python Tables/generate_tables.py; sources: Drafts/SCOPE_Tables.xlsx and Tables/published_baselines.json.
\begin{table}[htbp]
\centering
\setlength{\abovecaptionskip}{5pt}
\setlength{\belowcaptionskip}{6pt}
\caption{\textbf{LIBERO-PRO results.} Success rates (\%) under position swaps (Pos) and task changes (Task). Overall averages over suites, and All averages over both settings.}
\label{tab:libero-summary}
\small
\setlength{\tabcolsep}{1.6pt}
\renewcommand{\arraystretch}{1.16}
\begin{tabularx}{\linewidth}{@{}l*{9}{C}@{}}
\toprule
\multirow{2.5}{*}{\textbf{Method}} & \multicolumn{2}{c}{Object} & \multicolumn{2}{c}{Goal} & \multicolumn{2}{c}{Spatial} & \multicolumn{3}{c@{}}{Overall} \\
\cmidrule(lr){2-3}\cmidrule(lr){4-5}\cmidrule(lr){6-7}\cmidrule(lr){8-10}
 & Pos & Task & Pos & Task & Pos & Task & Pos & Task & All \\
\midrule
OpenVLA & $0.00$ & $0.00$ & $0.00$ & $0.00$ & $0.00$ & $0.00$ & $0.00$ & $0.00$ & $0.00$ \\
$\pi_0$ & $0.00$ & $0.00$ & $0.00$ & $0.00$ & $0.00$ & $0.00$ & $0.00$ & $0.00$ & $0.00$ \\
SUREFlow & $0.00$ & $0.00$ & $0.00$ & $0.00$ & $0.00$ & $0.00$ & $0.00$ & $0.00$ & $0.00$ \\
MolmoAct & $6.00$ & $0.00$ & $0.00$ & $0.00$ & $0.00$ & $0.00$ & $2.00$ & $0.00$ & $1.00$ \\
X-VLA & $2.00$ & $8.00$ & $1.00$ & $9.00$ & $0.00$ & $0.00$ & $1.00$ & $5.67$ & $3.33$ \\
$\pi_{0.5}$ & $17.00$ & $1.00$ & $38.00$ & $0.00$ & $20.00$ & $1.00$ & $25.00$ & $0.67$ & $12.83$ \\
CaP-Agent0 & $22.00$ & $18.00$ & $26.00$ & $17.00$ & $12.00$ & $14.00$ & $20.00$ & $16.33$ & $18.17$ \\
RATs & $61.00$ & $63.00$ & $43.00$ & $36.00$ & $29.00$ & $31.00$ & $44.33$ & $43.33$ & $43.83$ \\
RHO & $12.20$ & $37.20$ & $50.60$ & $55.60$ & $53.00$ & $61.60$ & $38.60$ & $51.47$ & $45.03$ \\
\midrule
\tablemethod{ASPIRE}{Opus 4.6} & $97.80$ & $95.40$ & $80.80$ & $45.40$ & $51.00$ & $59.60$ & $76.53$ & $66.80$ & $71.67$ \\
\addlinespace[3pt]
\tablemethod{ASPIRE}{GLM-5.3-Flash} & \tablehead{$98.07$\\[-1pt]{\scriptsize$\pm\,0.50$}} & \tablehead{$95.60$\\[-1pt]{\scriptsize$\pm\,3.82$}} & \tablehead{$83.40$\\[-1pt]{\scriptsize$\pm\,2.16$}} & \tablehead{$48.80$\\[-1pt]{\scriptsize$\pm\,11.68$}} & \tablehead{$54.47$\\[-1pt]{\scriptsize$\pm\,3.45$}} & \tablehead{$64.53$\\[-1pt]{\scriptsize$\pm\,1.17$}} & \tablehead{$78.64$\\[-1pt]{\scriptsize$\pm\,1.74$}} & \tablehead{$69.64$\\[-1pt]{\scriptsize$\pm\,4.78$}} & \tablehead{$74.14$\\[-1pt]{\scriptsize$\pm\,3.11$}} \\
\addlinespace[3pt]
\tablemethod{\textbf{SCOPE}}{GLM-5.3-Flash} & \tablehead{$\mathbf{99.40}$\\[-1pt]{\scriptsize$\pm\,0.60$}} & \tablehead{$\mathbf{97.73}$\\[-1pt]{\scriptsize$\pm\,0.61$}} & \tablehead{$\mathbf{88.33}$\\[-1pt]{\scriptsize$\pm\,0.61$}} & \tablehead{$\mathbf{61.87}$\\[-1pt]{\scriptsize$\pm\,5.41$}} & \tablehead{$\mathbf{65.40}$\\[-1pt]{\scriptsize$\pm\,3.82$}} & \tablehead{$\mathbf{71.20}$\\[-1pt]{\scriptsize$\pm\,1.11$}} & \tablehead{$\mathbf{84.38}$\\[-1pt]{\scriptsize$\pm\,1.25$}} & \tablehead{$\mathbf{76.93}$\\[-1pt]{\scriptsize$\pm\,2.19$}} & \tablehead{$\mathbf{80.66}$\\[-1pt]{\scriptsize$\pm\,1.46$}} \\
\bottomrule
\end{tabularx}
\par\vspace{4pt}
{\footnotesize\raggedright RHO and RATs use 50 and 10 evaluation trials per task, respectively.\par}
\end{table}

On Robosuite, SCOPE attains $84.48\,\pm\,1.67\%$ mean success
(Table~\ref{tab:robosuite}), improving over ASPIRE with the same
GLM-5.3-Flash backbone by 4.29 percentage points.
It improves all seven task means in the same-backbone
comparison. The largest gain is on two-arm lift, where success rises
from 70.33\% to 86.00\%.

% Generated by python Tables/generate_tables.py; sources: Drafts/SCOPE_Tables.xlsx and Tables/published_baselines.json.
\begin{table}[htbp]
\centering
\setlength{\abovecaptionskip}{5pt}
\setlength{\belowcaptionskip}{6pt}
\caption{\textbf{Robosuite results.} Success rates (\%) on seven manipulation tasks. Mean averages over tasks.}
\label{tab:robosuite}
\small
\setlength{\tabcolsep}{1.6pt}
\renewcommand{\arraystretch}{1.16}
\begin{tabularx}{\linewidth}{@{}l*{8}{C}@{}}
\toprule
\textbf{Method} & \tablehead{Cube\\lift} & \tablehead{Cube\\stack} & \tablehead{Cube\\restack} & \tablehead{Spill\\wipe} & \tablehead{Two arm\\handover} & \tablehead{Two arm\\lift} & \tablehead{Nut\\assembly} & \tablehead{Mean} \\
\midrule
CaP-Agent0 & $97.00$ & $98.00$ & $89.00$ & $\mathbf{100.00}$ & $20.00$ & $74.00$ & $0.00$ & $68.29$ \\
RHO$^{\dagger}$ & $98.00$ & $98.00$ & $54.00$ & $\mathbf{100.00}$ & $86.00$ & $53.00$ & $1.00$ & $70.00$ \\
\midrule
\tablemethod{ASPIRE}{Opus 4.6} & $97.00$ & $\mathbf{99.00}$ & $\mathbf{100.00}$ & $99.00$ & $\mathbf{92.00}$ & $71.00$ & $9.00$ & $81.00$ \\
\addlinespace[3pt]
\tablemethod{ASPIRE}{GLM-5.3-Flash} & \tablehead{$98.33$\\[-1pt]{\scriptsize$\pm\,1.53$}} & \tablehead{$97.33$\\[-1pt]{\scriptsize$\pm\,1.53$}} & \tablehead{$97.67$\\[-1pt]{\scriptsize$\pm\,1.53$}} & \tablehead{$98.33$\\[-1pt]{\scriptsize$\pm\,1.15$}} & \tablehead{$88.67$\\[-1pt]{\scriptsize$\pm\,5.13$}} & \tablehead{$70.33$\\[-1pt]{\scriptsize$\pm\,6.81$}} & \tablehead{$10.67$\\[-1pt]{\scriptsize$\pm\,4.73$}} & \tablehead{$80.19$\\[-1pt]{\scriptsize$\pm\,2.14$}} \\
\addlinespace[3pt]
\tablemethod{\textbf{SCOPE}}{GLM-5.3-Flash} & \tablehead{$\mathbf{100.00}$\\[-1pt]{\scriptsize$\pm\,0.00$}} & \tablehead{$\mathbf{99.00}$\\[-1pt]{\scriptsize$\pm\,1.00$}} & \tablehead{$\mathbf{100.00}$\\[-1pt]{\scriptsize$\pm\,0.00$}} & \tablehead{$\mathbf{100.00}$\\[-1pt]{\scriptsize$\pm\,0.00$}} & \tablehead{$91.67$\\[-1pt]{\scriptsize$\pm\,3.21$}} & \tablehead{$\mathbf{86.00}$\\[-1pt]{\scriptsize$\pm\,3.61$}} & \tablehead{$\mathbf{14.67}$\\[-1pt]{\scriptsize$\pm\,4.93$}} & \tablehead{$\mathbf{84.48}$\\[-1pt]{\scriptsize$\pm\,1.67$}} \\
\bottomrule
\end{tabularx}
\par\vspace{4pt}
{\footnotesize\raggedright $\dagger$ RHO's 100 evaluation trials per task include 10 development trial IDs.\par}
\end{table}

On the Quanta X2 robot from X Square Robot, we test base--arm
coordination (\emph{Cube into basket}), bimanual
coordination (\emph{Two arm lift}), and fine manipulation (\emph{Cube restack}).
Both methods use ten development placements and share the language model,
perception modules, and control interfaces. Each frozen program is tested
on 20 held-out placements. Hardware details, search budgets, and task sequences appear in
Appendix~\ref{sec:real-world-tasks} and Figure~\ref{fig:real-world-tasks}.
SCOPE achieves 66.7\% overall success versus ASPIRE's 40.0\%, improving
by 30, 20, and 30 percentage points on the three tasks, respectively
(Table~\ref{tab:real-world}). Cube restack remains the hardest task for
both methods, at 55.0\% and 25.0\% success.
During development, stateful full-trace replay checks recorded behavior
locally without requiring exact physical reconstruction of earlier scenes.

% Real-world trial counts supplied by the user.
% On 2026-09-24, the user corrected ASPIRE's Cube restack successes from 3 to 5.
% This correction takes precedence over the workbook, which may not yet be updated.
\begin{table}[htbp]
\centering
\setlength{\abovecaptionskip}{5pt}
\setlength{\belowcaptionskip}{6pt}
\caption{\textbf{Real-world generalization.} Successes/trials (success rate)
for frozen programs on 20 held-out physical configurations per task.
Overall pools 60 trials per method. Bold marks the higher success rate.}
\label{tab:real-world}
\small
\setlength{\tabcolsep}{5pt}
\renewcommand{\arraystretch}{1.18}
\begin{tabularx}{\linewidth}{@{}llCC@{}}
\toprule
\textbf{Task} & \textbf{Capability} & \textbf{ASPIRE} & \textbf{SCOPE} \\
\midrule
Cube into basket & Base--arm coordination & 9/20 (45.0\%) & \textbf{15/20 (75.0\%)} \\
Two arm lift & Bimanual coordination & 10/20 (50.0\%) & \textbf{14/20 (70.0\%)} \\
Cube restack & Fine manipulation & 5/20 (25.0\%) & \textbf{11/20 (55.0\%)} \\
\midrule
Overall & --- & 24/60 (40.0\%) & \textbf{40/60 (66.7\%)} \\
\bottomrule
\end{tabularx}
\end{table}

\begin{figure}[htb]
    \centering
    \includegraphics[width=\linewidth]{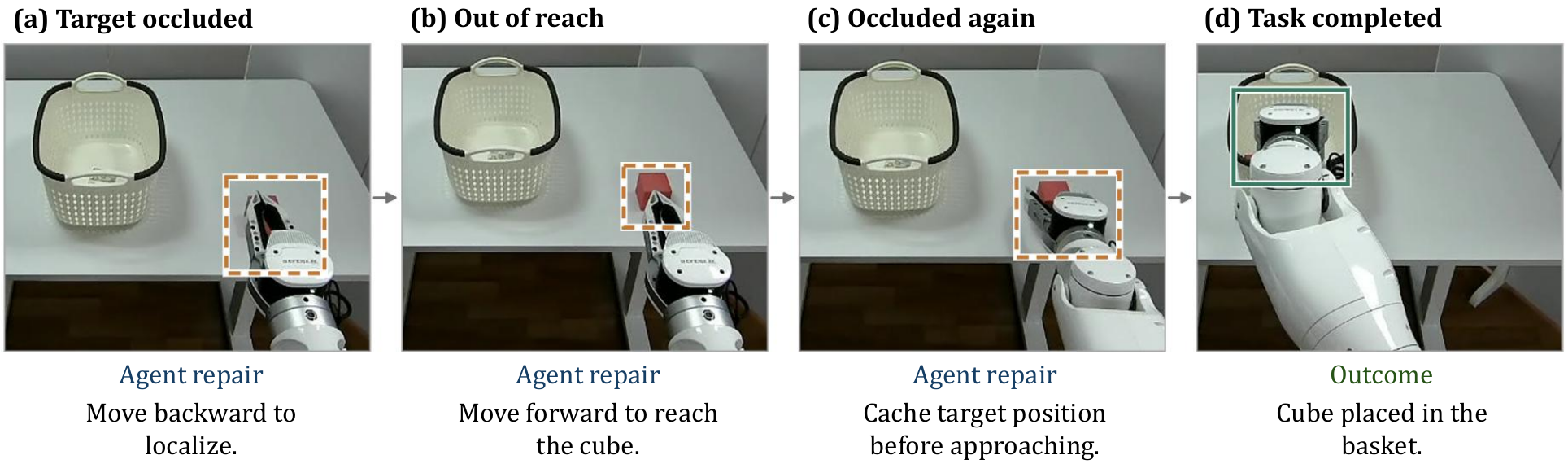}
    \caption{\textbf{Real-world repair of a cube-in-basket task.}
    \textbf{(a)} Gripper occlusion prompts backward motion for localization.
    \textbf{(b)} The target is visible but out of reach, prompting forward
    motion. \textbf{(c)} Renewed occlusion motivates caching the target
    position before approaching. \textbf{(d)} The revised program completes
    the task. Each panel pairs a key frame with the corresponding repair
    or outcome.}
    \label{fig:real-world-repair}
\end{figure}

Figure~\ref{fig:real-world-repair} presents a qualitative case of program
repair on a physical robot tasked with placing a cube into a basket.
The gripper initially occludes the cube, preventing localization.
Moving backward reveals the target but leaves it out of reach.
Moving forward restores reachability while introducing occlusion again.
The agent then revises the program to cache the target position before
approaching, enabling grasping and placement to proceed.

\FloatBarrier
\subsection{Component Ablations}
\label{sec:component-ablations}

We separately mask certification feedback, mask failure-frontier guidance,
or omit explicit revision-level selection on all ten tasks in each of
Goal--Task and Spatial--Pos. Variants share per-configuration repair limits and evaluation
protocols. Feedback masking leaves internal replay and environment
diagnostics active. Appendix~\ref{sec:ablation-protocol} specifies the
controls and retained information.

% Manually maintained ablation template; not overwritten by generate_tables.py.
% Full SCOPE: Drafts/SCOPE_Tables.xlsx, Table 2!E5 (Goal--Task)
% and Table 2!F5 (Spatial--Pos), scaled by 100 from the underlying run statistics.
% ASPIRE with GLM-5.3-Flash: Table 2!E4 and Table 2!F4, respectively.
% Variant boundaries for implementation:
% - Certification feedback: distinguish masking feedback from disabling replay;
%   disabling replay also requires a replacement for the diagnostic candidate set.
% - Frontier guidance: retain the internal graph needed for replay boundaries.
\begin{table}[htbp]
\centering
\setlength{\abovecaptionskip}{5pt}
\setlength{\belowcaptionskip}{6pt}
\caption{\textbf{Component ablations on LIBERO-PRO.} Success rates (\%) on all 10 tasks in each setting with GLM-5.3-Flash, reported as mean $\pm$ sample standard deviation over three runs.}
\label{tab:component-ablation}
\small
\setlength{\tabcolsep}{5pt}
\renewcommand{\arraystretch}{1.18}
\begin{tabularx}{\linewidth}{@{}lCC@{}}
\toprule
\textbf{Method / variant} & \textbf{Goal--Task} & \textbf{Spatial--Pos} \\
\midrule
ASPIRE (baseline) & $48.80\,\pm\,11.68$ & $54.47\,\pm\,3.45$ \\
\midrule
\textbf{Full SCOPE} & $61.87\,\pm\,5.41$ & $65.40\,\pm\,3.82$ \\
\addlinespace[3pt]
w/o certification feedback & $57.60\,\pm\,7.33$ & $59.33\,\pm\,4.16$ \\
w/o failure-frontier guidance & $55.07\,\pm\,9.10$ & $59.93\,\pm\,3.72$ \\
w/o explicit revision-level planning & $58.87\,\pm\,7.11$ & $62.13\,\pm\,3.88$ \\
\bottomrule
\end{tabularx}
\end{table}

All three removals lower mean success in both settings.
Removing failure-frontier guidance causes the largest decrease on
Goal--Task (6.80 percentage points), while removing certification feedback
causes the largest decrease on Spatial--Pos (6.07 points). These results
support combining failure localization, behavior-preservation feedback,
and explicit revision-level planning.

\subsection{Certification Rates and Costs}
\label{sec:certification-costs}

We measure how often replay certifies historical successes and the time
required to obtain this feedback (Figure~\ref{fig:certification}).
Table~\ref{tab:certification} in Appendix~\ref{sec:certification-details}
provides the full numerical results.
Certification rates count full matches when checking
older success witnesses under a different program, with unsupported
checks retained in the denominator. All quantities are task macro-averages.
Full matches account for 38.9--60.6\% of these checks across LIBERO-PRO
settings and 44.7\% on Robosuite. The remaining checks identify behavior
changes or unavailable replay evidence. Sparse environment diagnostics
assess the current program on selected unresolved configurations.

\begin{figure}[htb]
    \centering
    \includegraphics[width=\linewidth]{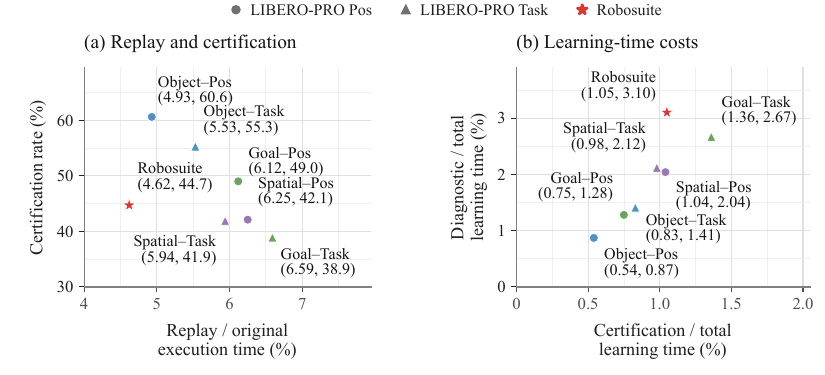}
    \caption{\textbf{Certification rates and feedback costs of SCOPE.}
    Each point is a task macro-average for one setting with GLM-5.3-Flash,
    covering 60 LIBERO-PRO task--setting pairs and seven Robosuite tasks.
    \textbf{(a)} Certification rate versus replay time relative to the
    corresponding original environment execution.
    \textbf{(b)} Certification-pipeline and diagnostic environment-execution
    times, each relative to total development learning time.
    Labels give $(x,y)$ in percent. Circles and triangles denote LIBERO-PRO
    Pos and Task settings, while stars denote Robosuite.
    The certification pipeline includes replay,
    retrieval, verification, and self-matching. Diagnostic time covers
    environment execution only.}
    \label{fig:certification}
\end{figure}

Replay takes 4.62--6.59\% of the corresponding original environment
execution time across the reported settings (Figure~\ref{fig:certification}a).
The full-trace certification
pipeline, including retrieval, verification, and
self-matching, accounts for 0.54--1.36\% of total development learning
time.
Diagnostic environment executions account for a further 0.87--3.10\%
of development time (Figure~\ref{fig:certification}b).
These measurements show that recorded executions
can provide behavioral evidence at a small fraction of their original
execution cost, while sparse diagnostics supply fresh observations for
further repair.

\FloatBarrier
\section{Conclusion and Limitations}
\label{sec:conclusion}

We presented SCOPE, which uses failed and successful executions as
complementary evidence for embodied heuristic learning.
Prerequisite-aware diagnosis guides revisions, while selective replay
assesses how accepted edits affect recorded successful behavior.
This feedback loop supports adaptation across configurations without
updating language-model parameters.
On LIBERO-PRO and Robosuite, SCOPE achieves mean held-out success rates
of 80.66\% and 84.48\%, exceeding the strongest baseline on each
benchmark by 6.51 and 3.48 percentage points, respectively.
Full-trace replay takes 4.62--6.59\% of the corresponding original
environment execution time, making historical behavior inexpensive
to assess during search.

Repeated language-model inference remains costly. Recorded token usage
implies an estimated cost of several thousand US dollars for running
the full experimental suite through hosted model APIs.
On physical robots, early candidate programs may produce unsafe behavior,
requiring human supervision and intervention.
These limitations motivate future work on more selective model querying
and runtime safety constraints to support efficient program learning
on physical robots.

\bibliography{references}
\bibliographystyle{iclr2027_conference}

\clearpage
\appendix
\section{Related Work}
\label{sec:related-work}

\paragraph{Embodied control and skill learning.}
Language models connect task instructions to embodied action through
grounded planning, multimodal representations, and structured control
interfaces~\citep{ichter2023saycan, driess2023palme, huang2025rekep}.
Code as Policies and ProgPrompt compose robot APIs into
programs with control flow and task
preconditions~\citep{liang2023code, singh2023progprompt}, while VoxPoser
constructs 3D value maps that express spatial affordances and constraints
for motion planning~\citep{huang2023voxposer}.
Environmental feedback and explicit world-state memory support adaptation
during execution in Inner Monologue and
Statler~\citep{huang2023inner, yoneda2024statler}.
Learning also extends to reusable skills. Voyager builds a library of
executable programs through an automatic curriculum in Minecraft, and
Playful Agentic Robot Learning acquires code skills through self-directed
robot exploration for subsequent
tasks~\citep{wang2024voyager, zhang2026playful}.
These representations support instruction grounding, state tracking, and
skill reuse. SCOPE makes task progress explicit in a diagnostic graph shared
across configurations. Prerequisite links connect execution evidence to
unresolved stages, allowing repairs to target causes beyond the visible
failure and select the revision level accordingly.

\paragraph{Embodied reinforcement learning.}
Embodied reinforcement learning and SCOPE share an iterative feedback
loop for improving behavior. An agent executes its current policy, uses
environment feedback to revise its behavior, and evaluates the update
through further interaction. QT-Opt learns visual grasping through
scalable off-policy Q-learning~\citep{kalashnikov2018qtopt}, while HIL-SERL combines
demonstrations and corrective human interventions with online reinforcement
learning for precise manipulation~\citep{luo2025hilserl}.
Reinforcement fine-tuning further adapts demonstration-trained policies
through interaction, including diffusion policies and vision-language-action
models~\citep{hu2025flare, ren2025dppo, chen2025conrft, li2026simplevlarl}.
Controlled evaluations show that reinforcement fine-tuning can
improve generalization under semantic and execution
shifts~\citep{liu2025rlvlageneralization}.
Reward synthesis and learned estimates of task progress enrich feedback
for policy learning and
planning~\citep{ma2023liv, ma2024eureka, shu2026rftf, shu2026feat2go}.
SCOPE implements this loop through successive control-program revisions,
carrying learned decision rules across trials while keeping language-model
parameters fixed. Its prerequisite-aware diagnoses
link feedback to local repairs, stage redesign, or strategy regeneration,
enabling targeted adaptation without policy-parameter training.
Under the assumptions in Section~\ref{sec:preliminary}, full-trace replay
also certifies success on matched recorded instances, making historical
interaction available as evidence of preserved behavior after an edit.

\paragraph{Feedback-driven program repair and search.}
When control is expressed as executable programs, feedback can guide code
revisions and the search over candidate policies.
Environment interaction, self-reflection, and tree search offer
complementary ways for language agents to revise their
decisions~\citep{yao2023react, shinn2023reflexion, zhou2024lats}.
Self-Refine iterates between self-feedback and revision, and Self-Debugging
combines program execution with natural-language explanations to repair
code~\citep{madaan2023selfrefine, chen2024selfdebug}.
Executable evaluation supports testing, candidate selection, and population
search over programs~\citep{chen2023codet, ni2023lever, romeraparedes2024funsearch}.
For robot control, CaP-Agent0 combines multi-turn visual feedback,
synthesized utilities, and parallel reasoning within
CaP-X~\citep{fu2026capx}.
ASPIRE is particularly relevant because it integrates per-primitive multimodal
traces, validated repair knowledge, and evolutionary search, evaluating
candidate programs through environment execution~\citep{lu2026aspire}.
RHO optimizes policy repositories using paired candidate evaluation and
a frontier of programs that maintains coverage of validation
instances~\citep{elmaaroufi2026rho}.
SCOPE augments this search with historical-success feedback after accepted
source updates. Stateful API replay identifies preserved traces and the
first divergent interactions, while sparse environment diagnostics provide
fresh evidence for repair. Subsequent revisions can therefore account for
an edit's effects on earlier successes as well as the current failure,
without repeating environment execution for certified traces.

\clearpage
\section{Experimental Setup Details}
\label{sec:experimental-details}

\paragraph{Benchmarks and evaluation protocol.}
We use LIBERO-PRO~\citep{zhou2025liberopro}, an extension of
LIBERO~\citep{liu2023libero}, and Robosuite~\citep{zhu2020robosuite}.
Following ASPIRE~\citep{lu2026aspire}, we evaluate the Object, Goal, and
Spatial suites under position swaps (Pos) and task changes (Task).
Each suite contains ten tasks, yielding 60 task--setting pairs.
We also evaluate seven Robosuite manipulation tasks, including object
lifting, stacking, wiping, bimanual manipulation, and nut assembly.
We retain ASPIRE's development and held-out seed splits, using seeds 51--65
and 1--50 for LIBERO-PRO, and seeds 101--125 and 1--100 for Robosuite.
Program generation, repair, and selection use the development split.
Within each run, we freeze one accepted program per LIBERO-PRO task--setting
pair and per Robosuite task. We then execute each program on every
corresponding held-out seed using current observations and control APIs,
without further program generation or repair. The execution archive and
critic are used only during learning.

\paragraph{Implementation and compute.}
We deploy GLM-5.3-Flash locally on eight NVIDIA PRO~6000 GPUs, using the
\texttt{max} reasoning setting and an agent sampling temperature of 1.0.
Perception services include Molmo2~\citep{clark2026molmo2},
SAM~3~\citep{carion2026sam3}, and GraspNet~\citep{fang2020graspnet}.
These services and simulation workloads run on a separate pool of eight
NVIDIA A100 GPUs.
Learning updates the control programs and execution memory while keeping
the language-model parameters fixed.

\paragraph{Simulation search and diagnostics.}
Each development configuration permits at most three main-search
repairs and one additional follow-up repair.
After each accepted source update, we sample at most $K=3$ historical success
witnesses for full-trace certification.
We then run at most $k_{\mathrm{diag}}=3$ diagnostic trials on uncertified
configurations, as described in Section~\ref{sec:certification-feedback}.

\paragraph{Comparison methods.}
We compare SCOPE with ASPIRE using the same GLM-5.3-Flash backbone and
include ASPIRE's Opus~4.6 results~\citep{lu2026aspire}.
Other program-based baselines include CaP-Agent0~\citep{fu2026capx},
RHO~\citep{elmaaroufi2026rho}, and RATs~\citep{zhang2026playful}.
For LIBERO-PRO, the vision-language-action baselines include
OpenVLA, $\pi_0$, and $\pi_{0.5}$ as reported by ASPIRE, together with
X-VLA~\citep{zheng2026xvla}, MolmoAct~\citep{lee2026molmoact}, and
SUREFlow~\citep{islam2026sureflow}.
We use official LIBERO-PRO leaderboard results~\citep{liberopro2026leaderboard}
for X-VLA and MolmoAct, and reported results for SUREFlow.
We compute aggregate success for the comparison methods as the equal-weight
mean of their reported suite--setting or task scores.
The other comparison methods retain their original training and inference protocols.
Table notes specify differences in evaluation trial counts and RHO's
Robosuite seed split.

\paragraph{Independent runs and reporting.}
For our GLM-5.3-Flash simulation experiments, we repeat the complete heuristic-learning
and held-out evaluation pipeline three times. Each repetition includes
fresh program generation, development, and evaluation of the resulting
frozen programs.
The development and held-out seed sets remain fixed across runs.
Success is determined by the benchmark task evaluator.
Within each run, we average success over held-out seeds and then over
tasks. LIBERO-PRO's overall score weights the six suite--setting
combinations equally. We report means and sample standard deviations
across the three run-level scores, capturing variation across complete
learning runs. Simulation benchmark and ablation results are percentages with two
decimal places. Results from the comparison methods' original reports are
point estimates, and bold indicates the best value or mean before rounding,
including ties.

\clearpage
\section{Ablation Protocol}
\label{sec:ablation-protocol}

The ablations in Table~\ref{tab:component-ablation} use GLM-5.3-Flash
on all ten tasks in each of Goal--Task and Spatial--Pos, the settings with
the largest gains over the same-backbone baseline.

\paragraph{Controlled comparisons.}
The four conditions use the same development and held-out configurations,
with three repetitions totaling 240 task-level learning runs. Each run
independently generates its initial program and task graph and maintains
its own archive and success bank. All use $K=3$, $k_{\mathrm{diag}}=3$,
and the same per-configuration main-search and follow-up repair limits.
Each ablation changes only the specified component. Acceptance
and frozen-program evaluation follow the same protocol. ASPIRE is shown
as a separate baseline.

\paragraph{Removing certification feedback.}
We mask full-trace certification results, first divergence details, and
derived summaries from planning and code generation. Internal replay and
the diagnostic and follow-up procedures remain unchanged, with observations
and failures from environment diagnostics still available.

\paragraph{Removing failure-frontier guidance.}
We mask the diagnostic graph, node labels, failure frontiers, evidence
windows, and derived summaries, while retaining raw observations, API
traces, errors, outcomes, and source programs.

\paragraph{Removing explicit revision-level planning.}
We retain failure diagnosis and repair planning but omit the explicit
choice among local repair, stage redesign, and strategy regeneration
before generating a complete candidate program.

\section{Detailed Certification Rates and Costs}
\label{sec:certification-details}

Table~\ref{tab:certification} provides the numerical results underlying
Figure~\ref{fig:certification}. The six LIBERO-PRO rows summarize ten tasks
each, and the Robosuite row summarizes seven tasks. All quantities are
task macro-averages. The time columns retain their respective denominators.

\begin{table}[htbp]
\centering
\setlength{\abovecaptionskip}{5pt}
\setlength{\belowcaptionskip}{6pt}
\caption{\textbf{Certification rates and costs of SCOPE.} Task macro-averages (\%) with GLM-5.3-Flash over 60 LIBERO-PRO task-setting pairs and seven Robosuite tasks.}
\label{tab:certification}
\small
\setlength{\tabcolsep}{3pt}
\renewcommand{\arraystretch}{1.18}
\begin{tabularx}{\linewidth}{@{}lc*{3}{C}@{}}
\toprule
\textbf{Setting} & \tablehead{\textbf{Certification}\\\textbf{rate}}
& \tablehead{\textbf{Replay time /}\\\textbf{original execution time}}
& \tablehead{\textbf{Certification time /}\\\textbf{total learning time}}
& \tablehead{\textbf{Diagnostic trial time /}\\\textbf{total learning time}} \\
\midrule
Object--Pos    & 60.6 & 4.93 & 0.54 & 0.87 \\
Object--Task   & 55.3 & 5.53 & 0.83 & 1.41 \\
\addlinespace[3pt]
Goal--Pos      & 49.0 & 6.12 & 0.75 & 1.28 \\
Goal--Task     & 38.9 & 6.59 & 1.36 & 2.67 \\
\addlinespace[3pt]
Spatial--Pos   & 42.1 & 6.25 & 1.04 & 2.04 \\
Spatial--Task  & 41.9 & 5.94 & 0.98 & 2.12 \\
\midrule
Robosuite      & 44.7 & 4.62 & 1.05 & 3.10 \\
\bottomrule
\end{tabularx}
\par\vspace{4pt}
{\footnotesize\raggedright
Diagnostic trial time covers environment execution only.
Total learning time includes model calls, environment executions, replay, and orchestration.
\par}
\end{table}

\clearpage
\section{Real-World Tasks and Protocol}
\label{sec:real-world-tasks}

We use the gripper-equipped Quanta X2 from X Square Robot, a wheeled
humanoid platform with two 7-degree-of-freedom (DoF) arms, a 4-DoF waist,
and a 2-DoF head~\citep{xsquareQuantax2SDK}. The platform has a 2-DoF
mobile base, stands 1.64\,m tall, and provides a nominal arm reach of
765\,mm~\citep{xsquareQuantax2}.

\begin{figure}[htb]
    \centering
    \includegraphics[width=\linewidth]{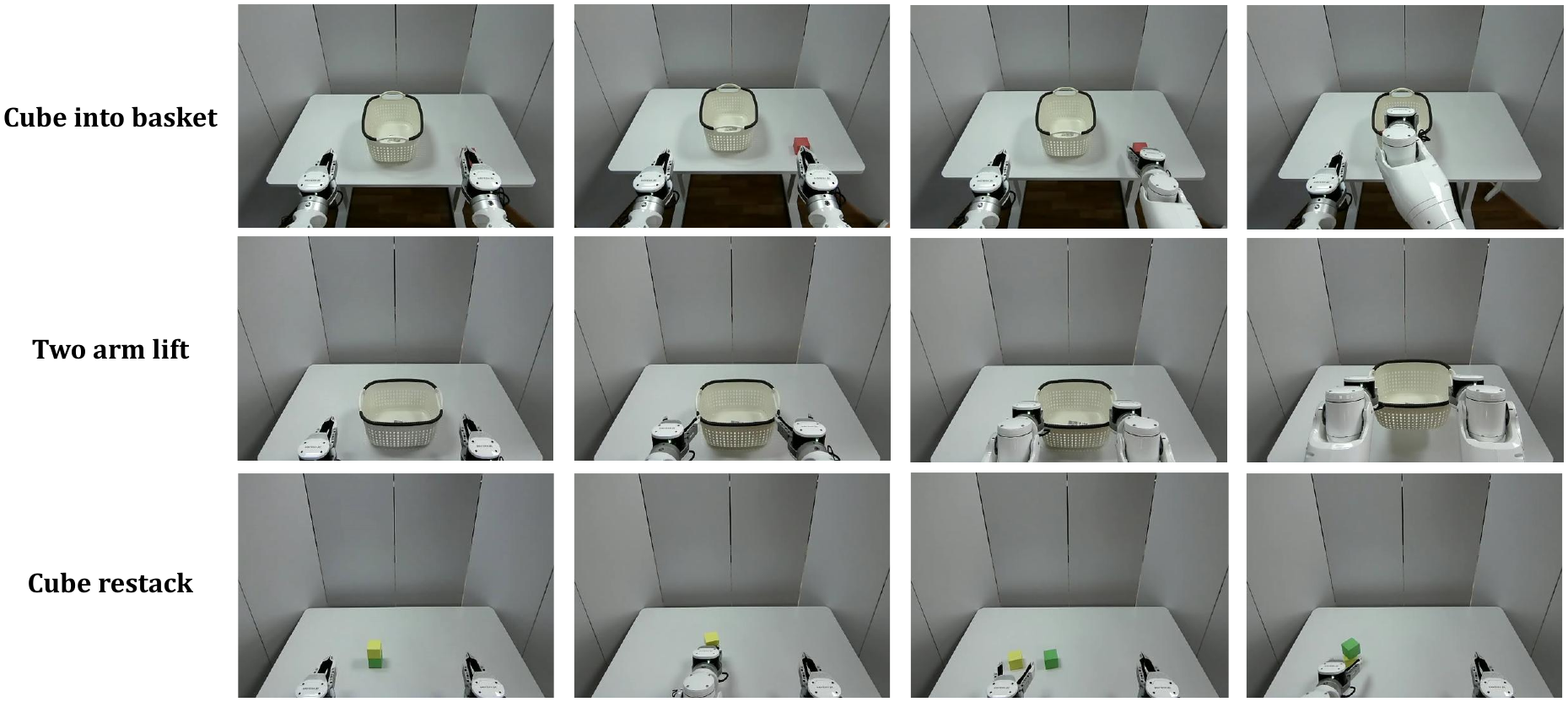}
    \caption{\textbf{Real-world task executions.}
    Representative frames progress from left to right for
    \emph{Cube into basket} (top), \emph{Two arm lift} (middle), and
    \emph{Cube restack} (bottom). The tasks evaluate mobile manipulation
    with base--arm coordination, bimanual coordination, and fine
    manipulation, respectively.}
    \label{fig:real-world-tasks}
\end{figure}

Figure~\ref{fig:real-world-tasks} illustrates the three physical tasks
evaluated in Table~\ref{tab:real-world}. The tasks cover base--arm
coordination, bimanual coordination, and fine manipulation, respectively.

Both methods use ten development configurations per task and share the
language model, perception modules, and robot control interfaces.
ASPIRE~\citep{lu2026aspire} uses five search rounds with eight candidate
programs per round. For each development configuration, SCOPE permits
at most three main-search repairs and one additional follow-up repair
after a failed physical diagnostic on a configuration left uncertified by
replay (Section~\ref{sec:certification-feedback}). After development,
we freeze one program per method and task and evaluate it on 20 held-out
initial placements without further repair.

\clearpage
\section{Results by LIBERO Task}
\label{sec:libero-task-results}

Tables~\ref{tab:libero-pos} and~\ref{tab:libero-task} report individual task
results for the Pos and Task settings, respectively, grouped by LIBERO-Object,
LIBERO-Goal, and LIBERO-Spatial.

\begin{table}[htbp]
\centering
\setlength{\abovecaptionskip}{5pt}
\setlength{\belowcaptionskip}{6pt}
\caption{Success rates (\%) for individual LIBERO tasks under the Pos setting.}
\label{tab:libero-pos}
\small
\setlength{\tabcolsep}{5pt}
\renewcommand{\arraystretch}{1.12}
\begin{tabularx}{\linewidth}{@{}lCCC@{}}
\toprule
Task & \tablehead{ASPIRE\\[-1pt]{\scriptsize Opus 4.6}} & \tablehead{ASPIRE\\[-1pt]{\scriptsize GLM-5.3-Flash}} & \tablehead{\textbf{SCOPE}\\[-1pt]{\scriptsize GLM-5.3-Flash}} \\
\midrule
\multicolumn{4}{@{}l}{\textit{LIBERO-Object}} \\
\addlinespace[2pt]
Alphabet soup & $\mathbf{100.00}$ & $98.67 \,\pm\, 1.15$ & $\mathbf{100.00} \,\pm\, 0.00$ \\
BBQ sauce & $\mathbf{100.00}$ & $99.33 \,\pm\, 1.15$ & $\mathbf{100.00} \,\pm\, 0.00$ \\
Butter & $\mathbf{100.00}$ & $98.67 \,\pm\, 2.31$ & $99.33 \,\pm\, 1.15$ \\
Chocolate pudding & $\mathbf{100.00}$ & $98.67 \,\pm\, 2.31$ & $\mathbf{100.00} \,\pm\, 0.00$ \\
Cream cheese & $\mathbf{100.00}$ & $98.00 \,\pm\, 2.00$ & $\mathbf{100.00} \,\pm\, 0.00$ \\
Ketchup & $92.00$ & $94.67 \,\pm\, 4.16$ & $\mathbf{96.67} \,\pm\, 3.06$ \\
Milk & $92.00$ & $96.67 \,\pm\, 4.16$ & $\mathbf{98.67} \,\pm\, 2.31$ \\
Orange juice & $\mathbf{100.00}$ & $98.67 \,\pm\, 1.15$ & $\mathbf{100.00} \,\pm\, 0.00$ \\
Salad dressing & $96.00$ & $97.33 \,\pm\, 1.15$ & $\mathbf{99.33} \,\pm\, 1.15$ \\
Tomato sauce & $98.00$ & $\mathbf{100.00} \,\pm\, 0.00$ & $\mathbf{100.00} \,\pm\, 0.00$ \\
\midrule
\multicolumn{4}{@{}l}{\textit{LIBERO-Goal}} \\
\addlinespace[2pt]
Open middle drawer & $72.00$ & $78.67 \,\pm\, 13.61$ & $\mathbf{89.33} \,\pm\, 4.16$ \\
Open top drawer + bowl & $86.00$ & $88.67 \,\pm\, 3.06$ & $\mathbf{98.67} \,\pm\, 1.15$ \\
Push plate to stove & $0.00$ & $2.00 \,\pm\, 0.00$ & $\mathbf{3.33} \,\pm\, 1.15$ \\
Bowl on plate & $92.00$ & $95.33 \,\pm\, 3.06$ & $\mathbf{100.00} \,\pm\, 0.00$ \\
Bowl on stove & $96.00$ & $98.67 \,\pm\, 1.15$ & $\mathbf{100.00} \,\pm\, 0.00$ \\
Bowl on cabinet top & $\mathbf{100.00}$ & $99.33 \,\pm\, 1.15$ & $\mathbf{100.00} \,\pm\, 0.00$ \\
Cream cheese in bowl & $74.00$ & $79.33 \,\pm\, 7.02$ & $\mathbf{92.67} \,\pm\, 3.06$ \\
Wine on rack & $98.00$ & $98.67 \,\pm\, 1.15$ & $\mathbf{100.00} \,\pm\, 0.00$ \\
Wine on cabinet top & $\mathbf{100.00}$ & $99.33 \,\pm\, 1.15$ & $\mathbf{100.00} \,\pm\, 0.00$ \\
Turn on stove & $90.00$ & $94.00 \,\pm\, 4.00$ & $\mathbf{99.33} \,\pm\, 1.15$ \\
\midrule
\multicolumn{4}{@{}l}{\textit{LIBERO-Spatial}} \\
\addlinespace[2pt]
Betw. plate + ramekin & $94.00$ & $96.67 \,\pm\, 2.31$ & $\mathbf{99.33} \,\pm\, 1.15$ \\
From table center & $32.00$ & $30.67 \,\pm\, 4.62$ & $\mathbf{48.67} \,\pm\, 13.01$ \\
In top drawer & $10.00$ & $20.00 \,\pm\, 19.08$ & $\mathbf{28.67} \,\pm\, 6.43$ \\
Next to cookie box & $16.00$ & $22.67 \,\pm\, 2.31$ & $\mathbf{48.67} \,\pm\, 6.43$ \\
Next to plate & $\mathbf{100.00}$ & $98.00 \,\pm\, 2.00$ & $99.33 \,\pm\, 1.15$ \\
Next to ramekin & $20.00$ & $13.33 \,\pm\, 8.08$ & $\mathbf{21.33} \,\pm\, 5.03$ \\
On cookie box & $24.00$ & $35.33 \,\pm\, 9.45$ & $\mathbf{54.67} \,\pm\, 2.31$ \\
On ramekin & $94.00$ & $94.00 \,\pm\, 4.00$ & $\mathbf{98.00} \,\pm\, 3.46$ \\
On stove & $86.00$ & $87.33 \,\pm\, 11.02$ & $\mathbf{96.00} \,\pm\, 3.46$ \\
On wooden cabinet & $34.00$ & $46.67 \,\pm\, 10.26$ & $\mathbf{59.33} \,\pm\, 13.32$ \\
\bottomrule
\end{tabularx}
\end{table}

\begin{table}[htbp]
\centering
\setlength{\abovecaptionskip}{5pt}
\setlength{\belowcaptionskip}{6pt}
\caption{Success rates (\%) for individual LIBERO tasks under the Task setting.}
\label{tab:libero-task}
\small
\setlength{\tabcolsep}{5pt}
\renewcommand{\arraystretch}{1.12}
\begin{tabularx}{\linewidth}{@{}lCCC@{}}
\toprule
Task & \tablehead{ASPIRE\\[-1pt]{\scriptsize Opus 4.6}} & \tablehead{ASPIRE\\[-1pt]{\scriptsize GLM-5.3-Flash}} & \tablehead{\textbf{SCOPE}\\[-1pt]{\scriptsize GLM-5.3-Flash}} \\
\midrule
\multicolumn{4}{@{}l}{\textit{LIBERO-Object}} \\
\addlinespace[2pt]
Alphabet soup & $\mathbf{100.00}$ & $98.00 \,\pm\, 2.00$ & $\mathbf{100.00} \,\pm\, 0.00$ \\
BBQ sauce & $\mathbf{100.00}$ & $99.33 \,\pm\, 1.15$ & $\mathbf{100.00} \,\pm\, 0.00$ \\
Butter & $98.00$ & $98.67 \,\pm\, 2.31$ & $\mathbf{100.00} \,\pm\, 0.00$ \\
Chocolate pudding & $\mathbf{100.00}$ & $98.67 \,\pm\, 1.15$ & $99.33 \,\pm\, 1.15$ \\
Cream cheese & $72.00$ & $84.00 \,\pm\, 5.29$ & $\mathbf{91.33} \,\pm\, 4.16$ \\
Ketchup & $\mathbf{96.00}$ & $88.00 \,\pm\, 19.08$ & $94.00 \,\pm\, 0.00$ \\
Milk & $\mathbf{100.00}$ & $98.67 \,\pm\, 1.15$ & $98.67 \,\pm\, 1.15$ \\
Orange juice & $\mathbf{100.00}$ & $99.33 \,\pm\, 1.15$ & $99.33 \,\pm\, 1.15$ \\
Salad dressing & $92.00$ & $94.00 \,\pm\, 5.29$ & $\mathbf{96.67} \,\pm\, 1.15$ \\
Tomato sauce & $96.00$ & $97.33 \,\pm\, 3.06$ & $\mathbf{98.00} \,\pm\, 2.00$ \\
\midrule
\multicolumn{4}{@{}l}{\textit{LIBERO-Goal}} \\
\addlinespace[2pt]
Open middle drawer & $0.00$ & $9.33 \,\pm\, 11.02$ & $\mathbf{30.00} \,\pm\, 10.00$ \\
Open top drawer + bowl & $46.00$ & $59.33 \,\pm\, 9.24$ & $\mathbf{80.00} \,\pm\, 6.93$ \\
Push plate to stove & $76.00$ & $67.33 \,\pm\, 20.53$ & $\mathbf{81.33} \,\pm\, 13.32$ \\
Bowl on plate & $88.00$ & $84.67 \,\pm\, 12.06$ & $\mathbf{90.67} \,\pm\, 9.45$ \\
Bowl on stove & $0.00$ & $8.67 \,\pm\, 8.33$ & $\mathbf{22.00} \,\pm\, 9.17$ \\
Bowl on cabinet top & $4.00$ & $7.33 \,\pm\, 5.77$ & $\mathbf{14.00} \,\pm\, 12.17$ \\
Cream cheese in bowl & $74.00$ & $72.67 \,\pm\, 22.48$ & $\mathbf{89.33} \,\pm\, 9.87$ \\
Wine on rack & $86.00$ & $94.00 \,\pm\, 4.00$ & $\mathbf{100.00} \,\pm\, 0.00$ \\
Wine on cabinet top & $14.00$ & $23.33 \,\pm\, 15.53$ & $\mathbf{36.00} \,\pm\, 17.32$ \\
Turn on stove & $66.00$ & $61.33 \,\pm\, 28.10$ & $\mathbf{75.33} \,\pm\, 5.77$ \\
\midrule
\multicolumn{4}{@{}l}{\textit{LIBERO-Spatial}} \\
\addlinespace[2pt]
Betw. plate + ramekin & $8.00$ & $12.00 \,\pm\, 8.00$ & $\mathbf{18.00} \,\pm\, 4.00$ \\
From table center & $64.00$ & $66.67 \,\pm\, 13.61$ & $\mathbf{80.00} \,\pm\, 9.17$ \\
In top drawer & $\mathbf{96.00}$ & $92.00 \,\pm\, 5.29$ & $90.67 \,\pm\, 10.07$ \\
Next to cookie box & $94.00$ & $96.67 \,\pm\, 1.15$ & $\mathbf{100.00} \,\pm\, 0.00$ \\
Next to plate & $70.00$ & $77.33 \,\pm\, 9.45$ & $\mathbf{82.00} \,\pm\, 8.72$ \\
Next to ramekin & $60.00$ & $\mathbf{64.00} \,\pm\, 24.25$ & $62.67 \,\pm\, 15.01$ \\
On cookie box & $70.00$ & $71.33 \,\pm\, 10.26$ & $\mathbf{92.00} \,\pm\, 5.29$ \\
On ramekin & $46.00$ & $61.33 \,\pm\, 13.32$ & $\mathbf{74.00} \,\pm\, 9.17$ \\
On stove & $42.00$ & $51.33 \,\pm\, 7.02$ & $\mathbf{58.00} \,\pm\, 6.00$ \\
On wooden cabinet & $46.00$ & $52.67 \,\pm\, 11.02$ & $\mathbf{54.67} \,\pm\, 6.11$ \\
\bottomrule
\end{tabularx}
\end{table}

\clearpage
\section{Behavioral Interference of Program Edits}
\label{sec:behavioral-interference}

Fix programs $P,P'$ and a task distribution $\mathcal D_d$. Compare their
executions on the same instance $\xi=(x_0,\omega,\kappa)$, sharing initial
conditions, randomness, and execution semantics. Under the assumptions in
Section~\ref{sec:preliminary}, identical external interactions, including
termination, yield the same binary task outcome. Let $E_{P,P'}$ denote the
event that these interactions differ, and define the edit's behavioral divergence probability
as $\eta(P,P')=\Pr(E_{P,P'})$. All probabilities and expectations below are
over $\xi\sim\mathcal D_d$.

\paragraph{Proposition (Behavioral divergence bounds).}
The change in expected success and the probability of losing a prior success
satisfy
\begin{equation}
\begin{aligned}
    J(P')-J(P)
      &=\mathbb E\bigl[(Y(P';\xi)-Y(P;\xi))\ind[E_{P,P'}]\bigr],\\
    |J(P')-J(P)|&\leq\eta(P,P'),\\
    \Pr\bigl(Y(P;\xi)=1,\,Y(P';\xi)=0\bigr)&\leq\eta(P,P').
\end{aligned}
\label{eq:behavioral-scope}
\end{equation}
The last probability measures regressions across all task instances.

\paragraph{Proof.}
Outside $E_{P,P'}$, the outcomes agree, so multiplying their difference by
$\ind[E_{P,P'}]$ leaves it unchanged. Taking expectations gives the first
identity. Since binary outcomes differ by at most one,
\[
    |J(P')-J(P)|
    \leq\mathbb E\bigl[|Y(P';\xi)-Y(P;\xi)|\ind[E_{P,P'}]\bigr]
    \leq\Pr(E_{P,P'}).
\]
The regression event $\{Y(P;\xi)=1,\,Y(P';\xi)=0\}$ is a subset of
$E_{P,P'}$, giving the final bound.

\paragraph{Edits within a conditional branch.}
Suppose $P$ and $P'$ differ only within a branch guarded by the same
side-effect-free predicate $g$. Let $q_P(g)$ be the probability that $P$
reaches this guard and enters the branch at least once during execution.
The programs execute identically before the first entry. If the branch is
never entered, the modified code is never executed. Thus,
\[
    E_{P,P'}\subseteq\{P\text{ enters the guarded branch at least once}\},
    \qquad \eta(P,P')\leq q_P(g).
\]

\paragraph{Implications for repair.}
An improvement $J(P')-J(P)=\Delta>0$ requires $\eta(P,P')\geq\Delta$.
A low behavioral divergence probability therefore limits the possible gain,
while the outcomes of changed executions determine whether performance
improves. This motivates broader revisions when local repairs are
insufficient. SCOPE selects revision levels from execution evidence and uses replay to
assess preservation on recorded instances.

\clearpage
\section{Certification Guarantee and Feedback Cost}
\label{sec:certification-analysis}

\subsection{Instance-Specific Certification}
\label{sec:instance-certification}

Fix a historical success witness $w=(P_w,\xi_w,T_w)$ with
$Y(P_w;\xi_w)=1$, and let $P$ be the program being checked.
The certificate $C_t(P,w)=1$ requires a complete match against the
stateful tape $T_w$ and normal termination, as defined in
Section~\ref{sec:certification-feedback}. Under the replay assumptions in
Section~\ref{sec:preliminary},
\begin{equation}
    C_t(P,w)=1\quad\Longrightarrow\quad Y(P;\xi_w)=1.
    \label{eq:certificate-soundness}
\end{equation}

\paragraph{Proof.}
Compare replay of the tested program $P$ against $T_w$ with its
execution in the environment on the original instance $\xi_w$.
They begin with the same program inputs and runtime semantics.
We induct over API events, maintaining agreement between their
program-visible values and reference relationships, and between the
real execution's external history and the witness tape.

Before the first event, the inputs agree and the external history is
empty. Suppose the correspondence holds before event $j$. Deterministic
program execution produces the same API request and pre-call public
state in both runs. Since replay matches, these equal
$(q_{w,j},B^-_{w,j})$. The real environment starts from the same instance
and has received the same preceding interactions as the witness.
Its response and public-state effects therefore match the recorded
$(v_{w,j},B^+_{w,j})$. Faithful replay reproduces the return value or
exception and the associated mutations, preserving shared references
and array views. The two runs of $P$ consequently agree after the event,
establishing the induction step.

A full match consumes every recorded event without additional external
interactions. Deterministic continuation then makes the real execution
terminate normally, as replay does. The resulting environment state
matches that of the successful witness, so the fixed evaluator gives
$Y(P;\xi_w)=Y(P_w;\xi_w)=1$.

The certificate applies to the recorded instance $\xi_w$. A fresh reset
of the same configuration may have different inputs or randomness.
A divergent or unavailable replay does not establish task failure.
Source acceptance remains determined by success on the current
configuration. Certification provides evidence for later repairs,
without requiring monotonic improvement in $J(P_t^{\mathrm{acc}})$ or
preservation of every historical success.

\subsection{Feedback Cost}
\label{sec:certification-cost-analysis}

Let $n_t^{\mathrm{tr}}$ and $n_t^{\mathrm{diag}}$ count selected replay
checks and diagnostic environment executions after search trial $t$.
The $T$ search trials include follow-up executions. Diagnostics are
counted separately. The per-update limits imply
$n_t^{\mathrm{tr}}\leq K A_t$ and
$n_t^{\mathrm{diag}}\leq k_{\mathrm{diag}}A_t$,
so both counts are zero when the accepted source is unchanged.
These limits give
\begin{equation}
\begin{aligned}
    Q_{\mathrm{tr}}
      &=\sum_{t=1}^{T}n_t^{\mathrm{tr}}
        \leq K\sum_{t=1}^{T}A_t,\\
    Q_{\mathrm{diag}}
      &=\sum_{t=1}^{T}n_t^{\mathrm{diag}}
        \leq k_{\mathrm{diag}}\sum_{t=1}^{T}A_t.
\end{aligned}
\label{eq:feedback-counts}
\end{equation}

Let $c_{\mathrm{tr}}$ and $c_{\mathrm{env}}$ denote the mean observed
costs of a selected replay check and a diagnostic environment execution
over the run. Their aggregate cost is
\begin{equation}
    C^{\mathrm{fb}}
      =Q_{\mathrm{tr}}c_{\mathrm{tr}}
       +Q_{\mathrm{diag}}c_{\mathrm{env}}.
    \label{eq:feedback-cost}
\end{equation}
The count bounds are independent of the success-bank size, while
replay time depends on trace length, the first divergence, and local
program execution. Total learning cost additionally includes search
executions, model queries, record processing, and orchestration.
Follow-up executions are included among the search trials. Their
diagnosis, planning, and generation requests contribute to model-query
cost. The certification timing reported in
Section~\ref{sec:certification-costs} also includes retrieval, record
verification, and self-matching in addition to the selected replay checks.

\section{AI Use Statement}
\label{sec:ai-use}

We used large language models to assist with language editing and polishing
of the manuscript. The authors take full responsibility for its content.

\end{document}